\documentclass{article}
\usepackage{bibspacing}
\usepackage{url}
\usepackage{spconf,amsmath,graphicx}
\usepackage{multirow}
\usepackage{makecell}
\usepackage{booktabs}
\usepackage{amssymb}
\usepackage{amsmath}
\usepackage{graphicx}
\usepackage{enumitem}
\usepackage{threeparttable}
\usepackage{diagbox}
\usepackage{setspace}
\usepackage{array,caption}
\usepackage[labelfont=bf]{caption}

\usepackage{hyperref}
\usepackage{color}

\title{$\text{M}^3$ST:~MIX AT THREE LEVELS FOR SPEECH TRANSLATION}
\name{Xuxin Cheng$^{1\dagger}$, 
Qianqian Dong$^2$, Fengpeng Yue$^2$, Tom Ko$^2$, Mingxuan Wang$^2$, Yuexian Zou$^{1,3*}$\thanks{
$\dagger$~Work was done while at ByteDance AI Lab.}
\thanks{*~Corresponding author: zouyx@pku.edu.cn.}
}
\address{$^1$ADSPLAB, School of ECE, Peking University, $^2$ByteDance AI Lab, $^3$Peng Cheng Laboratory\\
   chengxx@stu.pku.edu.cn \\ \{dongqianqian, yuefengpeng, tom.ko, wangmingxuan.89\}@bytedance.com, zouyx@pku.edu.cn}

\begin{document}
%
\maketitle
\begin{abstract}
How to solve the data scarcity problem for end-to-end speech-to-text translation~(ST)? It's well known that data augmentation is an efficient method to improve performance for many tasks by enlarging the dataset. In this paper, we propose \textbf{M}ix at three levels for \textbf{S}peech \textbf{T}ranslation~(M$^3$ST) method to increase the diversity of the augmented training corpus. Specifically, we conduct two phases of fine-tuning based on a pre-trained model using external machine translation~(MT) data. In the first stage of fine-tuning, we mix the training corpus at three levels, including word level, sentence level and frame level, and fine-tune the entire model with mixed data. At the second stage of fine-tuning, we take both original speech sequences and original text sequences in parallel into the model to fine-tune the network, and use Jensen-Shannon divergence to regularize their outputs. Experiments on MuST-C speech translation benchmark and analysis show M$^3$ST outperforms current strong baselines and achieves state-of-the-art results on eight directions with an average BLEU of 29.9.
\end{abstract}
\section{Introduction}
\vspace{-0.5em}
\label{sec:intro}
Speech translation~(ST), which has a wide range of applications, is a cross-modal task that the model receives acoustic speech signals and generates text translations in the target language. End-to-end ST system has attracted more and more attention and some work has demonstrated its superiority, even outperforming conventional cascaded systems\cite{RongYe2021EndtoendST,han2021learning}.

One of the major challenges in training an end-to-end ST model is the data scarcity. For example, there are only a few hundreds hours in MuST-C corpus whose size is far smaller than the data of automatic speech recognition~(ASR) or machine translation~(MT). How to make better use of limited labeled data and other parallel corpora from MT is promising but still a problem. Besides, data augmentation has been proven effective in alleviating this problem.

There are also some work investigating methods of data augmentation for ST, including DropDim\cite{zhang2022dropdim}, SkinAugment\cite{skinaugment}, STR\cite{lam-etal-2022-sample}, etc. However, existing work usually makes augmentations from a single level and has not been verified on strong baselines. Our work proposes a augmentation strategy in multiple levels, while demonstrating that there is a superposition of performance when used concurrently.
Inspired by MixSpeech\cite{meng2021mixspeech} for ASR, we introduce Mixup as data augmentation technique for ST and extend the original implementation.

In this paper, we propose M$^3$ST, an effective method for speech translation. In order to leverage limited data more effectively, we mix the training corpus at three levels, including word level, sentence level and frame level. At word level, we replace one noun from transcription by its similar word and perform the same operation in its corresponding speech and translation. At sentence level, we concatenate two sentences from different speakers. At frame level, we combine two different speech features with weight $\lambda$ as the input and use each target sequence to calculate the cross-entropy loss. Then we mix the two losses using the same weight $\lambda$. Experimental results show that our method achieves promising performance on the benchmark dataset MuST-C\cite{MattiaAntoninoDiGangi2019MuSTCAM}, and achieves state-of-the-art results on eight directions establishes whether using external MT data or not. 

\section{Method}
\vspace{-0.5em}

In this section, we first introduce the basic problem formulation of E2E-ST in Section 2.1 and show the overall model architecture in Section 2.2. Then, we introduce our proposed M$^3$ST in Section 2.3. Finally we introduce fine-tune with Jensen-Shannon divergence in Section 2.4.

\begin{figure*}[t]
  \centering
  \includegraphics[width=\linewidth]{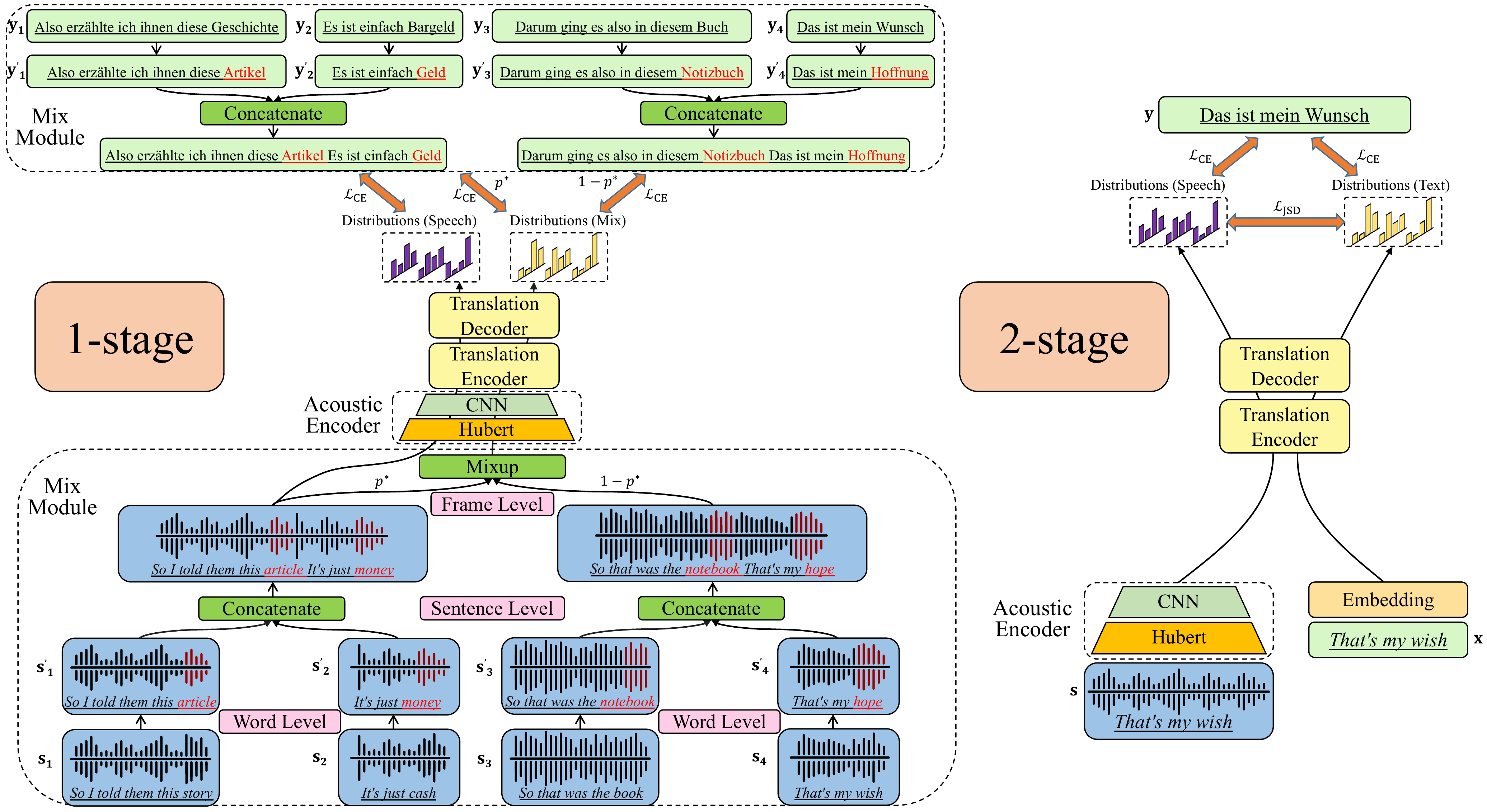}
  \caption{Overview of our proposed framework. In the 1-stage of fine-tuning, we mix the training corpus at three levels, including word level, sentence level and frame level, and fine-tune the entire model with mixed data. In the 2-stage of fine-tuning, both original speech sequence and transcription sequence are fed into the shared translation module to predict the translation, and we regularize two output predictions with an additional JS Divergence loss.}
  \label{fig:network}
\vspace{-1em}
\end{figure*}

\label{sec:pro}
\vspace{-1.25em}
\subsection{Problem Formulation}
\vspace{-0.75em}
The corpus of speech translation usually contains \textit{speech-transcription-translation} triples, which can be denoted as $\mathcal{D}=\{(\mathbf{s}, \mathbf{x}, \mathbf{y})\}$, where $\mathbf{s}$ is the sequence of audio wave, $\mathbf{x}$ is the corresponding transcription in the source language, and $\mathbf{y}$ is the translation in the target language. An end-to-end speech translation model aims to directly generate translation sequences $\mathbf{y}$ from speech sequences $\mathbf{s}$, without generating intermediate transcription $\mathbf{x}$. For the input sequence $\boldsymbol{\chi}$,  the standard training objective is to optimize the maximum likelihood estimation~(MLE) loss of the training data.
\begin{equation}
\setlength{\abovedisplayskip}{3pt}
\setlength{\belowdisplayskip}{3pt}
\mathcal{L}_{\mathrm{CE}}(\boldsymbol{\chi}, \mathbf{y})=-\sum_{i=1}^{|\mathbf{y}|} \log p_\theta\left(\mathbf{y}_i \mid \mathbf{y}_{<i}, \mathbf{h}(\boldsymbol{\chi})\right)
\end{equation}
\vspace{-1.25em}
\subsection{Model Architecture}
\vspace{-0.75em}
The overall structure of M$^3$ST is displayed in Fig.\ref{fig:network}, including  four modules: (a) a mix module~(see Section 2.3 for details) which mixes speech input and text input at three levels; (b) an acoustic encoder which extracts low-level features from acoustic input; (c) a translation encoder which extracts further semantic information from the output of acoustic encoder and word embedding; (d) a translation decoder which generates sentence tokens in the target language. 

\noindent \textbf{Acoustic Encoder}\ Recent work has shown that Hubert\cite{hsu2021hubert} has greater potential for improving the performance of end-to-end speech translation than Wav2vec2.0\cite{ye2022gigast}, so we use a pre-trained Hubert without any fine-tuning to extract acoustic representation. In addition, we add two convolutional layers to down sampling the extracted feature, since the length of the sequence of acoustic representations is much bigger than that of the sequence of sub-word embeddings.

\noindent \textbf{Translation Encoder}\ Our translation encoder follows the base configure, which is composed of 6 transformer\cite{AshishVaswani2022AttentionIA} encoder layers. Each layer includes a self-attention layer, a feed-forward layer, residual connections and normalization layers. The input of the translation encoder is the output of the acoustic encoder for ST task and the embedding of the transcription for MT task.

\noindent \textbf{Translation Decoder}\ Our translation decoder also follows the base configure, which is composed of 6 transformer\cite{AshishVaswani2022AttentionIA} decoder layers. Compared with a transformer encoder layer, a transformer decoder layer includes an additional cross-attention layer. The output of the translation encoder is fed here and used to predict the translation.

\noindent \textbf{Two Stages of Fine-tuning}\ We first pre-train the translation encoder and translation decoder using parallel \textit{transcription-translation} pairs from both the speech translation corpus and the external MT dataset. Then we fine-tune the entire model for the ST task using the mixed data in the first stage.  Finally we fine-tune the model with original ST data without mixing in the second stage.

\vspace{-1.25em}
\subsection{Mix Module}
\vspace{-0.75em}
Mixup technique\cite{HongyiZhang2017mixupBE} is applied in many tasks to improve the generalization limitation of empirical risk minimization. The main idea of Mixup is to improve the generalization ability of the model and enhance the robustness of the model. As we mentioned in Section 1, to leverage limited data more effectively, we present M$^3$ST to use mix at three levels, including word level, sentence level and frame level.

\noindent \textbf{Word Level}\ For a \textit{speech-transcription-translation} triple, denoted as  $\mathcal{D}=\{(\mathbf{s}, \mathbf{x}, \mathbf{y})\}$, where $\mathbf{s}=\left[\mathbf{s}_1, \mathbf{s}_2, \ldots, \mathbf{s}_{|\mathbf{s}|}\right]$, $\mathbf{x}=\left[\mathbf{x}_1, \mathbf{x}_2, \ldots, \mathbf{x}_{|\mathbf{x}|}\right]$, $\mathbf{y}=\left[\mathbf{y}_1, \mathbf{y}_2, \ldots, \mathbf{y}_{|\mathbf{y}|}\right]$, we first randomly choose a noun denoted $\mathbf{x}_i$ from $\mathbf{x}$ and one of its similar words denoted as $\mathbf{x}^{'}_i$. Then we get the corresponding speech $\mathbf{s}^{'}_i$ and translation $\mathbf{y}^{'}_i$ of $\mathbf{x}^{'}_i$. Finally we replace $\mathbf{s}_i$ by $\mathbf{s}^{'}_i$ and $\mathbf{y}_i$ by $\mathbf{y}^{'}_i$. Note that although we also use word-level alignment between speech and text, we use it extremely infrequently compared to STEMM\cite{QingkaiFang2022STEMMSW}.

\noindent \textbf{Sentence Level}\ For two \textit{speech-translation} pairs from different speakers, denoted as $(\mathbf{s}_i, \mathbf{x}_i, \mathbf{y}_i)$ and $(\mathbf{s}_j, \mathbf{x}_j, \mathbf{y}_j)$, we just concatenate them. Through this method, we can enhance the insensitivity of the model to the speaker and thus further improve the performance of the model.

\noindent \textbf{Frame Level}\ We set a combination weight, denoted as $\lambda$ and combine two different speech features with weight $\lambda$ as the input before they were fed into Hubert. To maintain a balanced mix, we ensure that every mixed pair with weight $\lambda$ will be mixed with weight $1-\lambda$. Then we use each target sequence to calculate the cross-entropy loss and mix the two losses using the same weight $\lambda$. 
\begin{equation}
\setlength{\abovedisplayskip}{3pt}
\setlength{\belowdisplayskip}{3pt}
    \begin{split}
    \textbf{s}_{mix}&={\lambda}\textbf{s}_i+(1-\lambda)\textbf{s}_j\\
    \mathcal{L}_{i}&=\mathcal{L}_{\rm{CE}}(\textbf{s}_{mix}, \textbf{y}_{i})\\
    \mathcal{L}_{j}&=\mathcal{L}_{\rm{CE}}(\textbf{s}_{mix}, \textbf{y}_{j})\\
    \mathcal{L}_{\rm{MIX}}(\bf{s}, \bf{y})&={\lambda}\mathcal{L}_{i}+(1-{\lambda})\mathcal{L}_{j}
    \end{split}
\end{equation}
\noindent where $\textbf{s}_{mix}$ is the mixed speech sequence and $\mathcal{L}_{\rm{MIX}}$ is the mix loss we propose. In the first stage of the fine-tuning, our training corpus consists of two parts, including original parallel \textit{speech-translation} pairs which more focus on the ST task itself and the mixed parallel \textit{speech-translation} pairs devoted to improve the generalization ability and enhance the robustness of the model. With the cross-entropy losses of two forward passes, the final training objective of the first stage of fine-tuning is as follows:
\begin{equation}
\setlength{\abovedisplayskip}{3pt}
\setlength{\belowdisplayskip}{3pt}
\mathcal{L}_{\mathrm{1}}(\boldsymbol{s}, \mathbf{y})=\mathcal{L}_{\mathrm{CE}}(\boldsymbol{s}, \mathbf{y})+\mathcal{L}_{\rm{MIX}}(\boldsymbol{s}, \mathbf{y})
\end{equation}
\vspace{-1.25em}
\subsection{Fine-tune With Jensen-Shannon Divergence}
\vspace{-0.75em}
In the second stage of fine-tuning, to enhance the knowledge transfer from MT task to ST task, we take both parallel \textit{speech-translation} pairs for ST task and \textit{transcription-translation} pairs for MT task as input. Besides, we introduce Jensen-Shannon Divergence (JSD) to regularize output predictions from the two tasks, which is
\begin{equation}
\setlength{\abovedisplayskip}{3pt}
\setlength{\belowdisplayskip}{3pt}
\begin{split}
\mathcal{L}_{\mathrm{JSD}}\left(\mathbf{s}, \mathbf{x}, \mathbf{y}\right)=\sum_{i=1}^{|\mathbf{y}|} \operatorname{JSD}\left\{p_\theta\left(\mathbf{y}_i \mid \mathbf{y}_{<i}, \mathbf{h}(\mathbf{s})\right) \|\right. \\
\left.p_\theta\left(\mathbf{y}_i \mid \mathbf{y}_{<i}, \mathbf{h}\left(\mathbf{x}\right)\right)\right\}
\end{split}
\end{equation}
where  $\mathbf{h}(\cdot)$  is the contextual representation output by the translation encoder, $p_{\theta}\left(\mathbf{y}_{i} \mid \mathbf{y}_{<i}, \mathbf{h}(\mathbf{s})\right)$  is the probability distribution of the \textit{i}-th target token predicted by the translation decoder of the speech sequence  $\mathbf{s}$, and  $p_{\theta}\left(\mathbf{y}_{i} \mid \mathbf{y}_{<i}, \mathbf{h}(\mathbf{x})\right)$  is that of the transcription sequence $\mathbf{x}$.
Our training loss of the second stage of fine-tuning consists of the following elements.
\begin{equation}
\setlength{\abovedisplayskip}{3pt}
\setlength{\belowdisplayskip}{3pt}
\mathcal{L}_{\mathrm{2}}(\boldsymbol{s}, \mathbf{y})=\mathcal{L}_{\mathrm{CE}}(\boldsymbol{s}, \mathbf{y})+\mathcal{L}_{\mathrm{CE}}(\boldsymbol{x}, \mathbf{y})+\mathcal{L}_{\mathrm{JSD}}\left(\mathbf{s}, \mathbf{x}, \mathbf{y}\right)
\end{equation}

\section{Experiments}
\label{sec:exp}
\vspace{-1.25em}
\subsection{Datasets}
\vspace{-0.75em}
\textbf{ST datasets}\ We conduct our experiments on all the translation directions in a multilingual speech translation corpus called MuST-C\cite{MattiaAntoninoDiGangi2019MuSTCAM}, which contains translations from English(En) to German(De), Spanish(Es), French(Fr), Italian(It), Dutch(Nl), Portuguese(Pt), Romanian(Ro) and Russian(Ru). Each direction includes a triplet of speech, transcription, and translation. As a benchmark ST dataset, the sizes of the eight directional TED talks range from 385 hours~(pt) to 504 hours~(es). 
In all experiments, model selection is based on the \texttt{dev} set and the final results are reported on the \texttt{tst-COMMON} set.

\begin{table*}[]
\setlength\tabcolsep{5pt}
\renewcommand{\arraystretch}{0.5}
\small
\centering
\caption{Case-sensitive detokenized BLEU scores on MuST-C tst-COMMON set. ``Speech" denotes unlabeled speech data. ``Text" means unlabeled text data. $\dagger$  use external 40M OpenSubtitles\cite{lison-tiedemann-2016-opensubtitles2016} MT data. * and ** mean the improvements over W2V2-Transformer baseline is statistically significant (p \textless 0.05 and p \textless 0.01, respectively).}
\begin{tabular}{l|cccc|cccccccc|c}
\midrule[1pt]
\multirow{2}{*}{\textbf{Models}} & \multicolumn{4}{c|}{\textbf{External Data}} & \multicolumn{9}{c}{\textbf{BLEU}}                                     \\
 &
  \multicolumn{1}{l}{\textbf{Speech}} &
  \multicolumn{1}{l}{\textbf{Text}} &
  \multicolumn{1}{l}{\textbf{ASR}} &
  \multicolumn{1}{l|}{\textbf{MT}} &
  \multicolumn{1}{l}{\textbf{De}} &
  \multicolumn{1}{l}{\textbf{Es}} &
  \multicolumn{1}{l}{\textbf{Fr}} &
  \multicolumn{1}{l}{\textbf{It}} &
  \multicolumn{1}{l}{\textbf{Nl}} &
  \multicolumn{1}{l}{\textbf{Pt}} &
  \multicolumn{1}{l}{\textbf{Ro}} &
  \multicolumn{1}{l}{\textbf{Ru}} &
  \multicolumn{1}{|l}{\textbf{Avg.}} \\
\midrule[1pt]
\multicolumn{14}{c}{\textit{Pre-train w/o external MT data}}                                                                                  \\
\midrule
Fairseq S2T$^{\prime}$20\cite{ChanghanWang2020fairseqSF}              & ×      & ×      & ×      & ×      & 22.7 & 27.2 & 32.9 & 22.7 & 27.3 & 28.1 & 21.9 & 15.3 & 24.8 \\
XSTNet$^{\prime}$21\cite{RongYe2021EndtoendST}                 & \checkmark      & ×      & ×      & ×      & 25.5 & 29.6  & 36.0 & 25.5 & 30.0 & 31.3 & 25.1 & 16.9 & 27.5 \\
TDA$^{\prime}$21\cite{xu2021stacked}                 & ×      & ×      & ×      & ×      & 25.4 & 29.6 & 36.1 & 25.1 & 29.6 & 31.1 & 23.9 & 16.4 & 27.2 \\
Revisit ST$^{\prime}$22\cite{zhang2022revisiting}              & ×      & ×      & ×      & ×      & 23.0 & 28.0 & 33.5 & 23.5 & 27.1 & 28.2 & 23.0 & 15.6 & 25.2 \\
STEMM$^{\prime}$22\cite{QingkaiFang2022STEMMSW}                   & \checkmark      & ×      & ×      & ×      & 25.6 & 30.3 & 36.1 & 25.6 & 30.1 & 31.0 & 24.3 & 17.1 & 27.5 \\
ConST$^{\prime}$22\cite{ye2022cross}                   & \checkmark      & ×      & ×      & ×      & 25.7 & 30.4 & 36.8 & 26.3 & 30.6 & 32.0 & 24.8 & 17.3 & 28.0 \\
W2V2-Transformer                   & \checkmark      & ×      & ×      & ×      & 24.3 & 29.6 & 35.2 & 25.1 & 29.1 & 30.3 & 23.4 & 16.5 & 26.7\\
M$^3$ST                   & \checkmark      & ×      & ×      & ×      & $\textbf{26.4}^{**}$ & $\textbf{31.0}^{**}$ & $\textbf{37.2}^{**}$ & $\textbf{26.6}^{**}$ & $\textbf{30.9}^{**}$ & $\textbf{32.8}^{**}$ & $\textbf{25.4}^{**}$ & $\textbf{18.3}^{**}$ & $\textbf{28.6}$\\
\midrule
\multicolumn{14}{c}{\textit{Pre-train w/ external MT data}}\\
\midrule
XSTNet$^{\prime}$21\cite{RongYe2021EndtoendST}         & \checkmark      & ×      & ×      &\checkmark      & 27.1 & 30.8 & 38.0 & 26.4 & 31.2 & 32.4 & 25.7 & 18.5 & 28.8 \\
SATE$^{\prime}$21\cite{xu2021stacked}                     & ×      & ×      & \checkmark      & \checkmark    & $28.1^{\dagger}$ & - & - & - & - & - & - & - & - \\
TaskAware$^{\prime}$21\cite{indurthi2021task}               & × & ×    & \checkmark   & \checkmark      & 28.9 & - & - & - & - & - & - & - & -  \\
STEMM$^{\prime}$22\cite{QingkaiFang2022STEMMSW}                   & \checkmark      & ×      & ×      & \checkmark     & 28.7 & 31.0 & 37.4 & 25.8 & 30.5 & 31.7 & 24.5 & 17.8 & 28.4 \\
ConST$^{\prime}$22\cite{ye2022cross}                   & \checkmark      & ×      & ×      & \checkmark      & 28.3 & 32.0 & 38.3 & 27.2 & 31.7 & 33.1 & 25.6 & 18.9 & 29.4 \\
W2V2-Transformer                   & \checkmark      & ×      & ×      & \checkmark     & 27.2 & 30.2 & 36.8 & 25.6 & 29.7 & 30.9 & 24.1 & 17.5 & 27.8\\
M$^3$ST                   & \checkmark      & ×      & ×      & \checkmark     & $\textbf{29.3}^{**}$ & $\textbf{32.4}^{**}$ & $\textbf{38.5}^{*}$ & $\textbf{27.5}^{**}$ & $\textbf{32.5}^{**}$ & $\textbf{33.4}^{**}$ & $\textbf{25.9}^{**}$ & $\textbf{19.3}^{**}$ & $\textbf{29.9}$\\
\midrule[1pt]
\end{tabular}
\label{tab:main_results}
\vspace{-2.0em}
\end{table*}

\noindent \textbf{MT datasets}\ We also use external MT datasets to pre-train our translation model, including WMT for En-De/Es/Fr/Ro/Ru directions, and OPUS100 datasets\cite{zhang2020improving}for En-It/Nl/Pt directions, as an expanded setup. For En-De, the size of WMT16 data is 4.6M.

\vspace{-1.25em}
\subsection{Experimental setups}
\vspace{-0.75em}
\textbf{Pre-processing}\ 
We use the raw 16-bit 16kHz mono-channel audio wave as speech input and remove the punctuation of the transcriptions. Both transcriptions and translations are case-sensitive. For each translation direction, a unigram SentencePiece\footnote{\label{v2}\url{https://github.com/google/sentencepiece}}\cite{TakuKudo2018SentencePieceAS} model is used to learn a vocabulary whose size is 10K on the text data from ST dataset. We use the vocabulary to partition the text in the ST and MT corpora into subword units. Considering the grammatical differences between languages, we only replace nouns with the practice mentioned above. We first collect words from transcriptions, then use CoreNLP\footnote{\label{v2}\url{https://downloads.cs.stanford.edu/nlp/software/stanford-corenlp-4.5.1.zip}} to split nouns from the words. To get word-level \textit{speech-transcription} alignment pairs, we use Montreal Forced Aligner\footnote{\label{v2}\url{https://github.com/MontrealCorpusTools/Montreal-Forced-Aligner}}\cite{mcauliffe2017montreal} toolkit. To get word-level \textit{transcription-translation} alignment pairs,we use fast\_align\cite{dyer2013simple} toolkit. We also use word2vec\cite{mikolov2013efficient} to search the the five closest words of each noun from the transcriptions. 

\noindent \textbf{Training Details}\ We use Adam optimizer\cite{kingma2014adam} to optimize the parameters in our model. We use EarlyStop which can stop training if the loss on the dev set didn't decrease within ten epochs, with no more than 16M source audio frames per batch. The mix ratio $\lambda$ is set to 0.4.

\noindent \textbf{Baseline Models}\ We compare our model with strong end-to-end ST models including:  XSTNet\cite{RongYe2021EndtoendST}, TaskAware\cite{indurthi2021task}, ConST\cite{ye2022cross}, STEMM\cite{QingkaiFang2022STEMMSW}, Fairseq ST\cite{ChanghanWang2020fairseqSF}, SATE\cite{xu2021stacked}, TDA\cite{du2022regularizing}, Revisit ST\cite{zhang2022revisiting}. We also implement a strong baseline W2V2-Transformer based on Hubert which has the same model architecture as our proposed M$^3$ST and is pre-trained in the same way, but we don't conduct the first stage of fine-tuning with the mix module.

\vspace{-1.25em}
\subsection{Main Results}
\vspace{-0.75em}

Table \ref{tab:main_results} shows experiment results of different models on MuST-C. Without external MT data, our model gains a large improvement of 0.6 BLEU (average over 8 directions) over the previous best models, and even outperforms W2V2-Transformer with external MT data except En-De, which proves the effectiveness of mix module. When external MT data is introduced, our method also outperforms SOTA by an average of 0.5 BLEU, which shows the potential of our M$^3$ST method.

\vspace{-1.0em}
\subsection{Analysis}
\vspace{-0.5em}
\subsubsection{Ablation Study}
\vspace{-0.5em}
To study the effectiveness of mixing at each level in our approach, we perform ablation experiments on En-De with external MT data, whose results are shown in Table \ref{tab:ablation}. We remove the mix at word level, at sentence level and at frame level in turn. From the results, we can observe the absence of mix at frame level leads to 0.5 BLEU drops, which is the largest of the three levels. This indicates that the mix at frame level can improve the performance of model significantly by introducing another speech, acting like a contrast signal to force the ST model to better translate the corresponding speech rather than being misled by other speech signals.
\renewcommand{\arraystretch}{0.6} 
\begin{table}[!ht]
\centering
\caption{Ablation results on En-De with external MT data.}
\vspace{-0.5em}
\begin{tabular}{lc}
\toprule[1pt]
\textbf{Models} & \textbf{BLEU} \\ \midrule[1pt]
M$^3$ST & 29.3 \\
\quad\textit{- word} & 29.1\\
\quad\textit{- word - sentence} & 28.8\\
\quad\textit{- word - sentence - frame} & 28.3\\
\bottomrule[1pt]
\end{tabular}
\vspace{-1em}
\label{tab:ablation}
\end{table}
\vspace{-1.0em}
\subsubsection{Effect of the Layer of Mixing at Frame Level}
\vspace{-0.5em}
We conduct a series of experiments to vary the layer of mixing at frame level, and the results are shown in Table \ref{tab:layer}. We can observe that the best BLEU can be obtained when we mix the speech at layer 0, i.e., when we mix the embedding of them. We suppose the reason for this is that more acoustic information has been lost when mixing at a deeper level.
\renewcommand{\arraystretch}{0.6} 
\begin{table}[!ht]
\centering
\caption{Results of mixing at different layer on En-De with external MT data.}
\vspace{-0.5em}
\begin{tabular}{cc}
\toprule[1pt]
\textbf{Layer} & \textbf{BLEU} \\ \midrule[1pt]
0 & 29.3 \\
2 & 29.1\\
4 & 28.6\\
6 & 28.3\\
\bottomrule[1pt]
\end{tabular}
\vspace{-1em}
\label{tab:layer}
\end{table}
\vspace{-1em}
\section{Conclusion}
\vspace{-0.5em}
In this paper, we propose the \textbf{M}ix at three levels for \textbf{S}peech \textbf{T}ranslation~(M$^3$ST) method to mix the training corpus at three levels, including word level, sentence level and frame level. Experiments and analysis demonstrate the effectiveness of our proposed method, which can leverage limited parallel pairs more efficiently. In the future, we will explore how to further solve the problem of data scarcity for ST task.
\section{Acknowledgements}
\vspace{-0.5em}
This paper was partially supported by Shenzhen Science \& Technology Research Program (No: GXWD2020123116580\\7007-20200814115301001) and NSFC (No: 62176008).

\bibliography{strings,refs}

\end{document}